\documentclass[runningheads]{llncs}
\usepackage[T1]{fontenc}
\usepackage{graphicx}
\usepackage{xfrac}
\usepackage{comment}
\usepackage{hyperref}
\usepackage{multirow}
\usepackage{todonotes}
\usepackage{fontawesome}
\usepackage{makecell}
\usepackage{tikz}
\def\checkmark{\tikz\fill[scale=0.15](0,.35) -- (.25,0) -- (1,.7) -- (.25,.15) -- cycle;} 

\begin{document}
\title{LLMs4OL 2024 Overview: The 1st Large Language Models for Ontology Learning Challenge}
\titlerunning{The 1st LLMs4OL Challenge @ ISWC 2024}
%
\author{
Hamed Babaei Giglou\orcidID{0000-0003-3758-1454} \and Jennifer D’Souza\orcidID{0000-0002-6616-9509} \and  S\"{o}ren Auer\orcidID{0000-0002-0698-2864}}
\authorrunning{Babaei Giglou et al.}
%
\institute{TIB Leibniz Information Centre for Science and Technology, Hannover, Germany\\
\email{\{hamed.babaei,jennifer.dsouza,auer\}@tib.eu}\\
}
\maketitle              

\begin{abstract}
This paper outlines the LLMs4OL 2024, the first edition of the Large Language Models for Ontology Learning Challenge. LLMs4OL is a community development initiative collocated with the 23rd International Semantic Web Conference (ISWC) to explore the potential of Large Language Models (LLMs) in Ontology Learning (OL), a vital process for enhancing the web with structured knowledge to improve interoperability. By leveraging LLMs, the challenge aims to advance understanding and innovation in OL, aligning with the goals of the Semantic Web to create a more intelligent and user-friendly web. In this paper, we give an overview of the 2024 edition of the LLMs4OL challenge\footnote{\href{https://sites.google.com/view/llms4ol}{https://sites.google.com/view/llms4ol}} and summarize the contributions.
\end{abstract}

\section{Introduction}
The Semantic Web aims to enrich the current web with structured knowledge and metadata, enabling enhanced interoperability and understanding across diverse systems. At the core of this endeavor is Ontology Learning (OL), a process that automates the extraction of structured knowledge from unstructured data~\cite{konys2019knowledge}, essential for constructing dynamic ontologies that underpin the Semantic Web. The emergence of Large Language Models (LLMs) like GPT-3~\cite{gpt3} and GPT-4~\cite{openai2024gpt4technicalreport} has revolutionized natural language processing (NLP), demonstrating remarkable performance across tasks such as language translation, question answering, and text generation. These models are particularly adept at crystallizing existing textual knowledge from a vast array of sources, making them potentially valuable for OL, where the goal is to extract a shared conceptualization of concepts and relationships from diverse inputs~\cite{gruber1995toward}. The introduction of LLMs has thus opened up new avenues of research, including the exploration of their potential in automating the OL process.

In our prior work published in the ISWC 2023 research track proceedings titled ``LLMs4OL: Large Language Models for Ontology Learning''~\cite{babaei2023llms4ol}, marked a notable direction towards employing LLMs in OL, demonstrating their potential in automating knowledge acquisition and representation for the Semantic Web. Based on this research, the \textbf{The 1st Large Language Models‌ for Ontology Learning Challenge at the 23rd ISWC 2024 (1st LLMs4OL Challenge @ ISWC 2024)} was introduced as a call for community development. With the LLMs4OL challenge, we aimed to catalyze community-wide engagement in validating and expanding the use of LLMs in OL. This initiative is poised to advance our comprehension of LLMs’ roles within the Semantic Web, encouraging innovation and collaboration in developing scalable and accurate ontology learning methods. 

LLMs4OL consists of three OL tasks, \textit{Task A -- Term Typing}, \textit{Task B -- Taxonomy Discovery}, and \textit{Task C -- Non-Taxonomic Relation Extraction}. While participation in all three tasks in the LLMs4OL 2024 challenge is stipulated as desirable, but not mandatory. Thus participants choose to enroll only in Task A or B or C, or Task A and B, or Task A and C, or Task B and C. Furthermore, participants are required to implement LLM-based solutions, we did not impose any restrictions on the LLM prompting methods. For instance, they can choose to bring in additional context information from the World Wide Web to enrich the training and test instances.  To thoroughly explore the potential of LLMs for OL, we structured the challenge around two distinct evaluation phases: \textit{(1) Few-shot testing phase} and \textit{(2) Zero-shot testing phase}. Through this work, we aim to contribute to the ongoing discourse on the capabilities of LLMs, particularly in the context of OL, and to provide insights into their potential for enhancing the Semantic Web. Thus, in the remainder of this paper, we detail the challenge tasks, what LLMs are being used, participant contributions, and findings.

\section{LLMs4OL 2024 Tasks}
\begin{figure*}[tb]
\includegraphics[width=\textwidth]{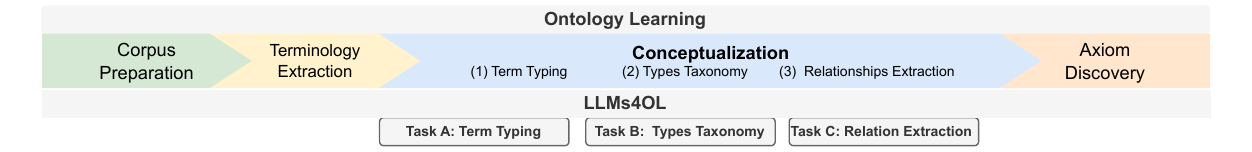}
\caption{The LLMs4OL task paradigm is an end-to-end framework for ontology learning. The three OL tasks that empirically validated in the LLMs4OL 2024 challenge, based on our prior research~\cite{babaei2023llms4ol}, are depicted within the blue arrow, aligned with the greater LLMs4OL paradigm.}\label{llms4ol}
\end{figure*}

In the LLMs4OL 2024 challenge, we have organized three main tasks which are‌ centered around the ontology primitives~\cite{maedche2001ontology} that comprise the following: \textbf{1.} a set of strings that describe terminological lexical entries $L$ for conceptual types; \textbf{2.} a set of conceptual types $T$; \textbf{3.} a taxonomy of types in a hierarchy $H_{T}$; \textbf{4.} a set of non-taxonomic relations $R$ described by their domain and range restrictions arranged in a heterarchy of relations $H_{R}$; and \textbf{5.} a set of axioms $A$ that describe additional constraints on the ontology and make implicit facts explicit. 

To address these primitives, the tasks for OL~\cite{noy2001ontology} are: 1) Corpus preparation -- collecting source texts for building ontology. 2) Terminology extraction -- extracting relevant terms from the texts. 3) Term typing -- grouping similar terms into conceptual types. 4) Taxonomy construction -- establishing ``is-a'' hierarchies between types. 5) Relationship extraction -- extracting semantic relationships beyond ``is-a'' between types. 6) Axiom discovery -- finding constraints rules for the ontology. These tasks constitute the LLMs4OL task paradigm as depicted in \autoref{llms4ol}. Assuming the corpus preparation step is done by reusing ontologies publicly released in the community, we introduced the following three main tasks for the first edition of the LLMs4OL challenge.

\begin{table}[t]
    \centering
    \caption{LLMs4OL 2024 challenge, subtasks, domains, number of participants per subtasks, and evaluation phases.}
    \label{tab:tasks}
    \begin{tabular}{|c|l|l|r|l|}
         \hline
         \textbf{\textit{Task}} & \textbf{\textit{SubTask}} & \textbf{\textit{Domain}} & \textbf{\textit{Participants}} & \textbf{\textit{Phase}} \\
         \hline
         \hline
         \multirow{10}{*}{A} & A.1 - WordNet & lexicosemantics & 7 & \multirow{9}{*}{Few-shot} \\
         \cline{2-4}
         &  A.2 - GeoNames & geographical locations & 5 & \\
         \cline{2-4}
         & A.3  - UMLS - NCI & \multirow{3}{*}{biomedical} & 5 &  \\
         &  A.3  - UMLS - MEDCIN &   & 4 & \\
         &  A.3  - UMLS - SNOMEDCT\_US &   & 4 & \\
         \cline{2-4}
         &  A.4  - GO - Biological Process & \multirow{3}{*}{biological} &  5 & \\
         &  A.4  - GO - Cellular Component &   & 5  &  \\
         &  A.4  - GO - Molecular Function &   & 5 &  \\
         \cline{2-5}
         &  A.5 - DBO & general knowledge & 2 &\multirow{2}{*}{Zero-shot} \\ 
         \cline{2-4}
         &  A.6 - FoodOn & food & 2 &   \\
         \hline
         \hline
         \multirow{6}{*}{B} &  B.1 - GeoNames & geographical locations & 5& \multirow{4}{*}{Few-shot} \\
         \cline{2-4}
         &  B.2 - Schema.org & web content types & 3 & \\
         \cline{2-4}
         &  B.3 - UMLS & biomedical &3& \\
         \cline{2-4}
         &  B.4 - GO & biological & 1 & \\
         \cline{2-5}
         &  B.5 - DBO & general knowledge &  2 & \multirow{2}{*}{Zero-shot} \\
         \cline{2-4}
          &  B.6 - FoodOn & food & 1 & \\
         \cline{2-4}
         \hline
         \hline 
         \multirow{3}{*}{C} &  C.1 - UMLS & biomedical & 2 & \multirow{2}{*}{Few-shot} \\
         \cline{2-4}
          &  C.2 - GO & biological &0& \\
         \cline{2-5}
          &  C.3 - FoodOn & food & 0 & \multirow{1}{*}{Zero-shot} \\
         \hline
    \end{tabular}
\end{table}

\subsection{Task A -- \underline{\textit{Term Typing}}} 
The \autoref{tab:tasks} shows 10 subtasks for \textit{Task A} across 6 distinct domains such as lexicosemantics, geographical locations,  biomedical, biological, general knowledge, and food domains. This task is defined as ''\textit{discover the generalized type for a given lexical term}''. For this task, for each ontology, participants are given training instances defined as following formalism. 
$$f_{prompt}^{TaskA}(L):= [S?].\;([L],\;[T])$$
Where $S$ is an optional context sentence (if available in the source ontology), $L$ is the lexical term prompted for, and $T$ is the conceptual term type. In the test phase, types are hidden, and participants predict them for given terms using their trained models. 

\subsection{Task B -- \underline{\textit{Taxonomy Discovery}}} 
After grouping terms under a conceptual type, in Task B, the goal is for given types ''\textit{discover the taxonomic hierarchy between types}'', where the hierarchy between types is defined with an "is-a" relationship. Participants receive training instances for 6 distinct subtasks (described in \autoref{tab:tasks}) as :
$$f_{prompt}^{TaskB}(a, b) := (T_a,\;T_b)$$
Where $T_a$ is the parent (superclass) of $T_b$, and $T_b$ is the child (subclass) of $T_a$. The goal is to train a system to correctly identify the taxonomy between type. The training dataset will include term types and taxonomically related type pairs. In the test phase, participants work with just term types and must use their trained models to identify correct taxonomic relationships (type pairs). The types for the training and test phases are mutually exclusive. Furthermore, for the testing phase participants are required to post-process their outputs to return type pairs that follow the order of superclass-subclass related types.

\subsection{Task C -- \underline{\textit{Non-Taxonomic Relation Extraction}}}  
Nonetheless, the "is-a" relations are not the only relations in ontologies. So, Task C aims to ''\textit{identify non-taxonomic, semantic relations between types}''. Training instances are given for three subtasks \textit{C.1 - UMLS}, \textit{C.2 - GO}, and  \textit{C.3 - FoodOn} as: 
$$f_{prompt}^{TaskC}(h, r, t):= (T_h, r, T_t)$$
Where,  $T_h$ and $T_t$ are head and tail taxonomic types, respectively, and $r$ is the non-taxonomic semantic relation between them, chosen from a predefined set $R$ of semantic relations. Participants aimed to train a system to identify pairs of types, and then classify pairs of types into semantic relations. The training phase involves types, relations, and triples of semantic relations; the test phase requires applying the trained system to predict semantically related triples from given types and the set of relations. 

The caveat here is that we do not expect participant systems to infer a semantic relation but rather establish semantically related types and classify their relation from a known set of predetermined relations. This implies that any manual ontology specification task predetermines which semantic relations hold for the given ontology. In an alternative scenario, where participants might have had to infer the semantic relation, we realize that the possibilities of semantic relations might have been rather vast. Hence we posit a more realistic task design by predetermining the possible set of semantic relations.  

\section{Evaluation}

There are two main evaluation phases for the challenge, which are the following:
\begin{itemize}
    \item \textbf{Few-shot testing phase.} Each ontology selected for system training will be divided into two parts: one part will be released for the training of the systems and another part will be reserved for the testing of systems in this phase. 
    \item \textbf{Zero-shot testing phase.}  New ontologies that are unseen during training will be introduced. The objective is to evaluate the generalizability and transferability of the LLMs developed in this challenge.   
\end{itemize}
For evaluation, we used the challenge datasets~\cite{llms4ol2024dataset} -- available at challenge GitHub\footnote{\href{https://github.com/HamedBabaei/LLMs4OL-Challenge-ISWC2024}{https://github.com/HamedBabaei/LLMs4OL-Challenge-ISWC2024}} repository -- with \textit{standard evaluation metrics} used for all tasks.  Given $\mathcal{G}(i)$ as a set of ground truth labels for sample $i$, and $\mathcal{P}(i)$ as a set of predicted labels for sample $i$, the precision $P$, recall $R$, and F1-score $F1$ are being calculated as follows:
\[
P = \frac{\sum_{i} |\mathcal{G}(i) \cap \mathcal{P}(i)|}{\sum_{i} |\mathcal{P}(i)|} \;,\;\; R = \frac{\sum_{i} |\mathcal{G}(i) \cap \mathcal{P}(i)|}{\sum_{i} |\mathcal{G}(i)|} \;,\;\; F1 = \frac{2 \times P \times R}{P + R}
\]
With precision, we assessed the percentage of the returned related pairs, while recall was used to measure the proportion of correct pairs that were accurately retrieved. In the end, the F1-score was calculated as the harmonic mean of precision and recall, serving as a comparison metric for the participants' submissions. We used Codalab\footnote{\href{https://codalab.lisn.upsaclay.fr/competitions/19547}{https://codalab.lisn.upsaclay.fr/competitions/19547}}~\cite{codalab} submission platform to organize participants submissions and scoring.
 
\section{Participant Systems and Results}

\begin{table}[t]
    \centering
    \tiny
    \caption{LLMs4OL 2024 challenge participants methods. $^*$ refers to the subtask that did not make the submission to the leaderboard but was reported in the paper. \texttt{MF} refers to "Molecular Function", \texttt{CC} refers to "Cellular Component", and \texttt{BF} refers to "Biological Process". \texttt{NCI}, \texttt{SNOMEDCT\_US}, and \texttt{MEDCIN} are from "UMLS".}
    \label{tab:comparison}
    \begin{tabular}{|c|c|c|p{2mm}|p{2mm}|p{2mm}|p{2mm}|p{2mm}|p{2mm}|p{2mm}|p{2mm}|p{2mm}|p{2mm}|p{2mm}|p{2mm}|p{2mm}|p{2mm}|p{2mm}|p{2mm}|p{2mm}|p{2mm}|p{2mm}|p{2mm}|}
        \hline
        \rotatebox{90}{\textbf{Team Name}} & \rotatebox{90}{\textbf{LLM of Use}} & \rotatebox{90}{\textbf{Approach}} & \rotatebox{90}{\textbf{Code}} & \rotatebox{90}{\textbf{A.1 - WordNet}}& \rotatebox{90}{\textbf{A.2 - GeoNames}} &  \rotatebox{90}{\textbf{A.3 - NCI}} &  \rotatebox{90}{\textbf{A.3 - MEDCIN}} &  \rotatebox{90}{\textbf{A.3 - SNOMEDCT\_US}} &  \rotatebox{90}{\textbf{A.4 - GO - BP}} &  \rotatebox{90}{\textbf{A.4 - GO - CC}} &  \rotatebox{90}{\textbf{A.4 - GO - MF}} &  \rotatebox{90}{\textbf{A.5 - DBO}} &  \rotatebox{90}{\textbf{A.6 - FoodOn}} &  \rotatebox{90}{\textbf{B.1 - GeoNames}} &  \rotatebox{90}{\textbf{B.2 - Schema.org}} &  \rotatebox{90}{\textbf{B.3 - UMLS}} &  \rotatebox{90}{\textbf{B.4 - GO}} &  \rotatebox{90}{\textbf{B.5 - DBO}} &  \rotatebox{90}{\textbf{B.6 - FoodOn}} &  \rotatebox{90}{\textbf{C.1 - UMLS}} &  \rotatebox{90}{\textbf{C.2 - GO}} &  \rotatebox{90}{\textbf{C.3 - FoodOn}}\\

        \hline
        \hline
        \textit{DSTI}~\cite{DSTI} & \makecell{Flan-T5\\GTE-Large} &  \makecell{Fine-tuning\\ RAG}& \href{https://github.com/HannaAbiAkl/SemanticTowers}{\faicon{github}} &  \checkmark & ${*}$ & & & & & & & &  & & & & & & & & &\\
        \hline
        \textit{DaSeLab}~\cite{DaSeLab} & \makecell{GPT-3.5-Turbo} & \makecell{Fine-tuning} & \href{https://github.com/AdritaBarua/LLMs4OL-2024-Task-A-Term-Typing}{\faicon{github}} &  \checkmark & \checkmark & \checkmark &\checkmark &\checkmark & & & & &  & & & & & & & & &\\
        \hline
        \textit{RWTH-DBIS}~\cite{RWTH-DBIS} &\makecell{GPT-3.5-Turbo\\LLaMA-3-8B}&\makecell{Prompting\\Fine-Tuning} &\href{https://github.com/MouYongli/LLMs4OL}{\faicon{github}}& \checkmark & \checkmark & \checkmark &\checkmark &\checkmark & \checkmark  & \checkmark  & \checkmark  & \checkmark & \checkmark  &\checkmark  &\checkmark  & & & & & & &\\
        \hline
        \textit{SKH-NLP}~\cite{SKH-NLP} & \makecell{LLaMA-3-70B\\Sentence-BERT}&\makecell{Prompting\\Fine-Tuning} &\href{https://github.com/s-m-hashemi/llms4ol-2024-challenge}{\faicon{github}}&   &   &  &  &  &   &  &  &  &  &\checkmark  &   &   & &  & & & &\\
        \hline
        \textit{TheGhost}~\cite{TheGhost} &\makecell{BLOOM-1B7\\BLOOM-3B\\BLOOM-
7B1\\LLaMA-7B\\LLaMA-2-7B\\LLaMA-3-8B\\BioMistral-7B\\OpenBioLLM-8B} & Prompt-Tuning& \href{https://github.com/themes12/Prompt-Tuning-for-LLMs4OL}{\faicon{github}}&\checkmark & \checkmark & \checkmark &\checkmark &\checkmark & \checkmark  & \checkmark  & \checkmark  &   &   &   &   & & & & & & &\\
        \hline
        \textit{silp\_nlp}~\cite{silpnlp} & \makecell{GPT-4o\\Mixtral-8x7B\\LLaMA-3-8B\\BERT\\Sentence-BERT}&\makecell{Prompting\\Fine-Tuning\\ML} &\href{https://drive.google.com/drive/folders/1vRynlNH6LouIvcI1ymHsm6DwYKSOUoAa?usp=sharing}{\faicon{github}}& \checkmark & \checkmark & \checkmark &\checkmark &\checkmark & \checkmark  & \checkmark  & \checkmark  & \checkmark & \checkmark  &\checkmark  &\checkmark  & \checkmark & &\checkmark & & \checkmark& &\\
        
        \hline
        \textit{Phoenixes}~\cite{Phoenixes} & \makecell{Mistral-7B\\Sentence-BERT}&\makecell{RAG} &\href{https://github.com/MahsaSanaei/Phoenixes-LLMs4OL-ISWC}{\faicon{github}}& \checkmark &   & \checkmark & &  & \checkmark  & \checkmark  & \checkmark  &   &   &\checkmark  &\checkmark  & \checkmark &\checkmark &\checkmark &\checkmark & \checkmark& &\\
        \hline
        \textit{TSOTSALearning}~\cite{TSOTSALearning} & \makecell{GPT-4\\BERT} & \makecell{RAG\\Rules} &‌\href{https://github.com/sudo-001/LLMs4OL-2024}{\faicon{github}}& \checkmark &  \checkmark  &  &  &  & \checkmark  & \checkmark  & \checkmark  &   &   &\checkmark  &  &   &  &  & & & &  \\
        \hline
    \end{tabular}
\end{table}

The LLMs4OL 2024 challenge has inspired diverse solutions, showcasing the growing potential of LLMs for OL tasks. Using the Codalab submissions platform, for this challenge we set a limit of 10 submissions per day and a total of 30 submissions per subtask. We received 272 total submissions from 14 participants. In final, this challenge attracted the interest of the final eight research teams, as demonstrated by the various approaches they submitted for the subtasks. Each subtask of the competition depicted a rigorous field inherent to OL, which helped facilitate breakthroughs in finding generalized types (Task A), identifying taxonomic hierarchies (Task B), and extracting non-taxonomic relations (Task C), further scaffolding future AI advancements. Notably, teams employed varied strategies to tackle subtasks, such as fine-tuning, prompt-tuning, and retrieval-augmented generation (RAG). These approaches were used to analyze OL tasks across domains like lexicosemantics, geographical locations, biomedical concepts, and more (see \autoref{tab:tasks} for subtasks and domains involved in this challenge). The summary of explored LLMs and subtasks are presented in \autoref{tab:comparison} and in the following we will detail contributions and findings.

\subsection{Participants Contributions}
The results for Task A are presented in \autoref{task-a-results}, for Task B in \autoref{task-b-results}, and for Task C in \autoref{task-c-results}. 

\begin{table}[!htb]
\centering
\caption{Task A - Term Typing Results for SubTasks}
    \label{task-a-results}
    \begin{tabular}{|l|l|r|r|r|}
    \hline
    \textbf{SubTask} & \textbf{Team Name} & \textbf{F1-Score} & \textbf{Precision} & \textbf{Recall} \\
    \hline
    \multirow{7}{*}{A.1 (FS) - WordNet} & TSOTSALearning & \textbf{0.9938} & \textbf{0.9938} & \textbf{0.9938} \\
    & DSTI & 0.9716 & 0.9716 & 0.9716 \\
    & DaseLab & 0.9697 & 0.9689 & 0.9704 \\
    & RWTH-DBIS & 0.9446 & 0.9446 & 0.9446 \\
    & TheGhost & 0.9392 & 0.9389 & 0.9395 \\
    & Silp\_nlp & 0.9037 & 0.9037 & 0.9037 \\
    & Phoenixes & 0.8158 & 0.7689 & 0.8687 \\
    \hline
    \multirow{5}{*}{A.2 (FS) - GeoNames} & DaseLab&  \textbf{0.5906} &  0.5906 &  \textbf{0.5906} \\
    & Silp\_nlp & 0.4433 & \textbf{0.7503} & 0.3146 \\
    & RWTH-DBIS & 0.4355 & 0.4355 & 0.4355 \\
    & TSOTSALearning & 0.2937 & 0.2937 & 0.2937 \\
    & TheGhost & 0.1489 & 0.1461 & 0.1519 \\
    \hline
    \multirow{5}{*}{A.3 (FS) - UMLS - NCI} & DaseLab&  \textbf{0.8249} & 0.8161 &  \textbf{0.8340}\\
    & Silp\_nlp & 0.6974 &  \textbf{0.8792} & 0.5779 \\
    & TheGhost & 0.5370 & 0.4450 & 0.6769 \\
    & RWTH-DBIS & 0.1691 & 0.1821 & 0.1579 \\
    & Phoenixes & 0.0737 & 0.0562 & 0.1070 \\
    \hline
    \multirow{4}{*}{A.3 (FS) - UMLS - MEDCIN }& Silp\_nlp & \textbf{0.9382} & \textbf{0.9591} & 0.9181 \\
    & DaseLab & 0.9373 & 0.9379 & \textbf{0.9366} \\
    & TheGhost & 0.5328 & 0.4183 & 0.7336 \\
    & RWTH-DBIS & 0.4566 & 0.4607 & 0.4526 \\
    \hline
    \multirow{4}{*}{ A.3 (FS) - UMLS - SNOMEDCT\_US} & DaseLab & \textbf{0.8829} & \textbf{0.8810}& \textbf{0.8848} \\
    & Silp\_nlp & 0.7552 & 0.8583 & 0.6742 \\
    & TheGhost & 0.5275 & 0.4266 & 0.6910 \\
    & RWTH-DBIS & 0.4747 & 0.4888 & 0.4613 \\
    \hline
    \multirow{5}{*}{A.4 (FS) - GO - Cellular Component}& Silp\_nlp & \textbf{0.2726} & \textbf{0.4279} & 0.2000 \\
    & RWTH-DBIS & 0.2178 & 0.1846 & \textbf{0.2656} \\
    & TheGhost & 0.1877 & 0.1653 & 0.2171 \\
    & TSOTSALearning & 0.0638 & 0.0767 & 0.0545 \\
    & Phoenixes & 0.0158 & 0.0124 & 0.0217 \\
    \hline
    \multirow{5}{*}{A.4 (FS) - GO - Biological Process} & Silp\_nlp & \textbf{0.2691} & \textbf{0.4006} & \textbf{0.2026} \\
    & TheGhost & 0.1025 & 0.0964 & 0.1095 \\
    & RWTH-DBIS & 0.0881 & 0.0693 & 0.1207 \\
    & TSOTSALearning & 0.0648 & 0.0806 & 0.0542 \\
    & Phoenixes & 0.0319 & 0.0214 & 0.0622 \\
    \hline
    \multirow{5}{*}{A.4 (FS) - GO - Molecular Function} & Silp\_nlp & \textbf{0.2970} & \textbf{0.4185} & \textbf{0.2302} \\
    & RWTH-DBIS & 0.1418 & 0.1670 & 0.1231 \\
    & TheGhost & 0.1270 & 0.1278 & 0.1261 \\
    & TSOTSALearning & 0.0910 & 0.1072 & 0.0790 \\
    & Phoenixes & 0.0700 & 0.0485 & 0.1256 \\
    \hline
    \multirow{2}{*}{A.5 (ZS) - DBO }& RWTH-DBIS & \textbf{0.4270} & \textbf{0.4270} & \textbf{0.4270} \\
    & Silp\_nlp & 0.3009 & 0.3009 & 0.3009 \\
    \hline
    \multirow{2}{*}{A.6 (ZS) - FoodOn} & RWTH-DBIS & \textbf{0.8068} & \textbf{0.8068} & \textbf{0.8068} \\
    & Silp\_nlp & 0.7278 & 0.7278 & 0.7278 \\
    \hline
    \end{tabular}
\end{table}
\noindent\textbf{DSTI~\cite{DSTI}. }DSTI fine-tuned Flan-T5-Small~\cite{chung2022scalinginstructionfinetunedlanguagemodels} model for \textit{SubTasks A.1 - WordNet} and \textit{A.2 - GeoNames}. Obtained F1-score of 0.9716 for \textit{SubTask A.1} and ranked as a second team. But for GeoNames they did not submit the model to the leaderboard due to the larger nature of GeoNames dataset that required more computational resources. They introduced two approaches for OL. The first approach is fine-tuning LLMs using the zero-shot prompting method, the second approach is using a RAG pipeline using the General Text Embeddings (GTE)-Large~\cite{li2023generaltextembeddingsmultistage} model as a retriever and fine-tuned LLM as a retriever. Due to the computational resources they preferred to use the Flan-T5-small model, and the results showed the effectiveness of their approach. 

\noindent\textbf{RWTH-DBIS~\cite{RWTH-DBIS}.} This team participated in tasks A and B (12 subtasks in total). For both tasks, they proposed a domain-specific continual training, fine-tuning, and knowledge-enhanced prompt-tuning approach. The models are firstly enriched with conceptual information related to terms and types. This is followed by CausalLM manner and task-specific fine-tuning using LLaMA-3-8B~\cite{dubey2024llama3herdmodels}. The proposed approach performs well on several subtasks, showcasing that incorporating domain-specific information and providing a list of classification types enhances inference performance. They concluded that in Task A, GPT-3.5-Turbo~\cite{openai2024gpt3.5} outperformed fine-tuned open-source LLM, and incorporating domain-specific information and providing a list of types at prompt significantly enhances the performance.

\noindent\textbf{DaSeLab~\cite{DaSeLab}.} The DaSeLab team participated in \textit{UMLS, GeoNames, and WordNet} subtasks. This team approach is based on fine-tuning a GPT-3.5-Turbo model. The result of fine-tuning on \textit{UMLS} and \textit{GeoNames} domains showed that fine-tuning of such model can achieve superior performance. The DaSeLab ranked first place in \textit{NCI} (0.8249), \textit{GeoNames} (0.5906), and \textit{SNOMEDCT\_US} (0.8829) subtasks (scores inside practices are F1-scores).

\noindent\textbf{TheGhost~\cite{TheGhost}.} The TheGhost team investigated a variety of LLMs with a prompt-tuning approach. They are the first team in the challenge that explored 11 LLMs (the LLM list depicted in \autoref{tab:comparison}) for 8 subtasks of term typing tasks within a few-shot testing evaluation scenario. They showed the viability of soft prompt tuning for OL and the challenge of imbalanced class prompt tuning. Their finding supports the complexity of geographical and biological domains at the term typing task of OL.

\noindent\textbf{silp\_nlp~\cite{silpnlp}.} The silp\_nlp team participated in all three tasks with a total of 15 subtasks. They ranked in first place in several subtasks including \textit{A.3 (FS) - UMLS - MEDCIN} ( 0.9382), \textit{A.4 (FS) - GO - Cellular Component} (0.2726), \textit{A.4 (FS) - GO - Biological Process} (0.2691), \textit{A.4 (FS) - GO - Molecular Function} (0.2970), \textit{B.2 (FS) - Schema.org} (0.6157), \textit{B.3 (FS) - UMLS}, \textit{B.5 (FS) - DBO} (0.2109), and \textit{C.1 (FS) - UMLS} (0.0783). They employed several machine learning techniques, such as Random Forest, Logistic Regression, and XGBoost, alongside advanced generative models like LLaMA-3-8B, Mixtral~\cite{jiang2024mixtralexperts}, and GPT-4o~\cite{openai2024gpt4technicalreport}. The results revealed that prompt-based methods were effective in some domains but not universally applicable. Notably, Random Forest models excelled in subtasks A.1 through A.4, while GPT-4o dominated the zero-shot tasks A.5 and A.6, as well as relation extraction tasks B and C. This team obtained in first-place in six subtasks and second place in five subtasks.

\noindent\textbf{TSOTSALearning~\cite{TSOTSALearning}.} The TSOTSALearning team focused on LLMs such as BERT~\cite{devlin-etal-2019-bert} and GPT-4. Through experimentation on \textit{SubTask A.1 - WordNet} dataset, they achieved an F1-score of 0.9264 with GPT-4, but significantly improved results when they combined BERT with rule-based strategies, leading to an F1-score of 0.9938 and ranked first place in \textit{WordNet} dataset. Their findings showed the importance of incorporating rules into LLMs for enhanced accuracy in OL. However, they highlight the challenge of identifying appropriate rules, suggesting that future work should focus on automating rule detection and integrating them seamlessly into LLMs. The \textit{WordNet} dataset is being considered as a low number of types and having a higher number of types makes it challenging to obtain highly accurate rules.

\noindent\textbf{SKH-NLP~\cite{SKH-NLP}.} Team SKH-NLP participated in \textit{SubTask B.1 - GeoNames}, where they developed a fine-tuning approach using the LLaMA-3-70B and BERT-Large~\cite{devlin2019bertpretrainingdeepbidirectional}. This team obtained the first place in \textit{SubTask B.1 - GeoNames} with an F1-score of 0.6557. Their comprehensive analysis demonstrates that BERT-Large, when fine-tuned, achieves performance close to the larger LLaMA-3-70B model. 

\noindent\textbf{Phoenixes~\cite{Phoenixes}.} The Phoenixes team explored the application of a Retrieval Augmented Generation (RAG) approach within the 12 subtaks of the challenge. They introduced a promising RAG-specific formulation over all three tasks of OL, where a RAG system with minor changes was developed for both tasks A and B, later can be used as a two-step approach for task C. Task C consists of the following steps: Step 1 -- runs the Task B approach for finding child-parent pairs and step 2 -- applying the Task A approach for assigning the relations to the pairs. They incorporated Mistral-7B~\cite{jiang2023mistral7b} as LLM and Dense Passage Retrieval (DPR)~\cite{karpukhin2020densepassageretrievalopendomain} model as the retriever model in the RAG framework. However, their results in both zero-shot and few-shot fall shorter than the fine-tuned models and this suggests that still fine-tuning is the key to obtain a high performance within OL.

\begin{table}[t]
    \centering
    \caption{Task B - Taxonomy Discovery Results for SubTasks}
    \label{task-c-results}
    \begin{tabular}{|l|l|r|r|r|}
    \hline
    \textbf{SubTask} & \textbf{Team Name} & \textbf{F1-Score} & \textbf{Precision} & \textbf{Recall} \\
    \hline
    \multirow{5}{*}{B.1 (FS) - GeoNames} & SKH-NLP & \textbf{0.6557} & \textbf{0.6318} & \textbf{0.6814} \\
    & RWTH-DBIS & 0.3409 & 0.2400 & 0.5882 \\
    & Silp\_nlp & 0.0830 & 0.0446 & 0.5931 \\
    & TSOTSALearning & 0.0104 & 0.0052 & 0.5294 \\
    & Phoenixes & 0.0036 & 0.0019 & 0.0294 \\
    \hline
    \multirow{3}{*}{B.2 (FS) - Schema.org }& Silp\_nlp & \textbf{0.6157} & 0.4578 & \textbf{ 0.9396} \\
    & RWTH-DBIS & 0.5733 & \textbf{0.5475} & 0.6016 \\
    & Phoenixes & 0.0155 & 0.0079 & 0.3901 \\
    \hline
    \multirow{3}{*}{B.3 (FS) - UMLS} & Silp\_nlp & \textbf{0.3544} & \textbf{0.4118} & 0.3111 \\
    & Phoenixes & 0.0960 & 0.0550 & 0.3778 \\
    & RWTH-DBIS & 0.0491 & 0.0257 & \textbf{0.5556} \\
    \hline
    B.4 (FS) - Gene Ontology (GO) & Phoenixes & 0.0164 & 0.0180 & 0.0149 \\
    \hline
    \multirow{2}{*}{B.5 (FS) - DBpedia Ontology (DPO)} & Silp\_nlp & \textbf{0.2109} & 0.1412 & 0.4164 \\
    & Phoenixes & 0.0164 & 0.0180 & 0.0149 \\
    \hline
    B.6 (ZS) - Food Ontology (FoodOn) & Phoenixes & 0.0308 & 0.0243 & 0.0420 \\
    \hline
\end{tabular}
\end{table}

\begin{table}[t]
    \centering
    \caption{Task C - Non-Taxonomic Relation Extraction Results for SubTasks}
    \label{task-b-results}
    \begin{tabular}{|l|l|r|r|r|}
    \hline
    \textbf{SubTask} & \textbf{Team Name} & \textbf{F1-Score} & \textbf{Precision} & \textbf{Recall} \\
    \hline
    \multirow{2}{*}{C.1 (FS) - UMLS} & Silp\_nlp & 0.0783 & \textbf{0.0494} & \textbf{0.1888} \\
    & Phoenixes & 0.0273 & 0.0433 & 0.0199 \\
    \hline
    \end{tabular}
\end{table}
\subsection{Large Language Models}
The participants in the challenge utilized a diverse array of LLMs, each bringing distinct strengths to the tasks. We detailed a breakdown of the key strengths of the prominent LLMs used.

\noindent\textsc{\textbf{gpt family}} -- GPT-3.5-Turbo, GPT-4, and GPT-4o: GPT based LLMs, developed by OpenAI, are renowned for their advanced natural language understanding and generation capabilities. These models excel in context comprehension and can handle a variety of queries effectively, making them particularly suitable for tasks that require deep semantic understanding and detailed generation. Their ability to generalize from a wide range of training data allows them to perform well across various knowledge domains relevant ontologies~\cite{babaei2023llms4ol,giglou2024llms4ommatchingontologieslarge}. GPT-3.5-Turbo was a popular choice among participants, with teams such as DaSeLab, RWTH-DBIS, and silp\_nlp using the model and demonstrating its high adaptability and effectiveness across the various challenge subtasks. Furthermore, GPT-4 and GPT-4o as more advanced models over GPT-3, were explored by the teams: TSOTSA Learning and \textit{silp\_nlp}.
    
\noindent\textsc{\textbf{llama family}} -- LLaMA-7B, LLaMA-2-7B, LLaMA-3-8B, and LLaMA-3-70B: The LLaMA models were another prominent choice among participants. With models like LLaMA-2 and LLaMA-3 featured by TheGhost, RWTH-DBIS, SKH-NLP, and silp\_nlp, their popularity stems from their open-source, efficiency, and scalability. These models' strengths in handling large-scale data and intricate details made them well-suited for comprehensive multi-dimensional data interpretation.
    
\noindent\textsc{\textbf{bloom family}} -- BLOOM-1B7, BLOOM-3B, and BLOOM-7B1: BLOOM~\cite{workshop2023bloom176bparameteropenaccessmultilingual} models, featured in our original research work \cite{babaei2023llms4ol}, gained traction due to their open-access nature and collaborative development. TheGhost, in particular, utilized a range of BLOOM models for their flexibility and multilingual capabilities. 

\noindent\textsc{\textbf{biomedical family}} -- BioMistral-7B and OpenBioLLM-8B: BioMistral-7B~\cite{labrak2024biomistral}, as a domain-specific fine-tuned variant of Mistral-7B, and OpenBioLLM-8B~\cite{OpenBioLLMs}, as a domain-specific fine-tuned variant of LLaMA-3-8B, were utilized for their domain-specific strengths in biomedical contexts. TheGhost's use of these models highlights their importance in tasks requiring detailed biomedical terminology and concepts, emphasizing their significance in the specialized subfields of the challenge.

\noindent\textsc{\textbf{mistral family}} -- Mistral-7B and Mixtral-8x7B: Mistral-7B, part of the Mistral family of models, was noted for its performance in the challenge by teams like Phoenixes and TheGhost. Moreover, Mixtral-8x7B was utilized by the team silp\_nlp.

\noindent\textsc{\textbf{others}} -- Flan-T5, GTE-Large, Sentence-BERT, and DPR: Flan-T5 and GTE-Large were chosen for their adaptability and fine-tuning capabilities. DSTI recognized their potential in fine-tuning and handling diverse NLP tasks efficiently when there are limited computational resources. Sentence-BERT was prominently used for tasks involving semantic similarity and sentence-level embeddings. Its popularity among participants like SKH-NLP and Phoenixes.  Phoenixes used DPR for the retrieval model of the RAG approach. 

\subsection{Trade-offs Between Precision and Recall}
Across the tasks, a clear trend emerges among the participating teams. Teams like silp\_nlp often exhibit high precision but lower recall, particularly in subtasks related to GO and UMLS ontologies. This suggests that while silp\_nlp is adept at avoiding false positives and making accurate predictions, it frequently misses relevant instances, indicating a more conservative approach. However, teams such as RWTH-DBIS and Phoenixes display a different trend, where recall is relatively higher than precision. These teams retrieve a larger number of relevant results but at the cost of precision, indicating that they tend to capture a broad set of possible answers, including many false positives. Their approach may be useful in tasks where coverage is prioritized over accuracy, but it also introduces challenges in filtering out noise.

Teams that manage to balance both precision and recall, such as DaSeLab and SKH-NLP, stand out for their well-rounded performance. These teams perform consistently across different subtasks by finding a middle ground between retrieving enough relevant results and minimizing false positives. DaSeLab, for example, shows balanced performance across multiple subtasks, especially in UMLS-related tasks, suggesting a more effective strategy. Meanwhile, SKH-NLP stands out in the GeoNames taxonomy discovery task, where it achieves high precision and recall, demonstrating its capability to capture relevant information without sacrificing accuracy.

In more challenging tasks, such as non-taxonomic relation extraction, the disparity between precision and recall becomes particularly pronounced. For example, both silp\_nlp and Phoenixes struggle, with silp\_nlp showing low precision but managing to retrieve more relevant results than Phoenixes, which has very low recall. This suggests that these tasks may require more sophisticated models or techniques to achieve higher performance. Overall, the results reflect that teams vary significantly in how they prioritize precision and recall, depending on the specific subtask, with some teams excelling in precision-oriented tasks while others show better results in recall-sensitive subtasks.

\section{Discussion}
\noindent\textbf{Performance Analysis.} As the participating teams navigated through the zero-shot and few-shot testing phases of the LLMs4OL 2024 challenge, notable variations in performance underscored the importance of model adaptability and data-specific adjustments. Few-shot tasks, particularly those involving geographical, biological, and biomedical domains, highlighted the critical need for specialized model tuning and the strategic use of training data to achieve high precision and recall rates. This indicates that achieving optimal performance in real-world ontology challenges requires not only selecting the right LLMs but also fine-tuning them to align with the specific characteristics of the domains and tasks at hand. Additionally, studies show that for Task A, even smaller models like Flan-T5-Small with 80M parameters can perform well when there are fewer types. However, as the number of types increases, larger models, such as those with 7B parameters, tend to perform better. One reason for the popularity of 7B models is that Parameter-Efficient Fine-Tuning (PEFT)~\cite{xu2023parameterefficientfinetuningmethodspretrained} fine-tuning requires less memory compared to traditional fine-tuning methods. Many participants also incorporated external knowledge, such as type definitions, synthesis data using LLMs, or general knowledge graphs (KGs) to build answer sets. These strategies have demonstrated a positive impact on fine-tuning performance.

\noindent\textbf{Complexity Across Domains and Tasks. }The results indicated that certain domains and tasks, such as biomedical term typing and non-taxonomic relation extraction, were more challenging than others. The variation in performance across tasks, particularly in relation to term complexity (e.g., Gene Ontology), highlights the complexity of certain knowledge domains. This still requires specialized approaches. The Phoenixes (on all three tasks) and DSTI (on task A only) teams introduced a formulation based on Retrieval-Augmented Generation (RAG) approaches with success, indicating that combining LLM generation capabilities with retrieval mechanisms can enhance accuracy in OL tasks. This approach is particularly suitable due to the hybrid framework with high adaptability to be extended with different components.

\noindent\textbf{Few-Shot and Zero-Shot Testing Phases. }While many models performed well in the few-shot phase, the zero-shot testing phase exposed limitations in the generalization capabilities of LLMs. Models like GPT-3.5 and GPT-4 demonstrated strong performance, but there were notable drops when transitioning from few-shot to zero-shot testing phases. More research is needed to improve the transferability and robustness of LLMs across unseen domains and ontologies.

\noindent\textbf{Task A vs Task C. }From a task perspective, Task C attracted only two teams, indicating it was perceived as highly challenging. Non-taxonomic relation extraction requires identifying complex relationships between terms that go beyond hierarchical (taxonomy-based) relations, which is a significantly more intricate task. Unlike simple is-a relationships, non-taxonomic relations are more diverse, context-dependent, and require a deeper understanding of the subject matter. Extracting these relations often involves dealing with ambiguous or implicit connections, requiring models to infer meanings that might not be explicit. This complexity might have discouraged more teams from participating, as success in this task requires advanced techniques, often combining deep semantic understanding with domain-specific knowledge. On the other hand, Task A, term typing, had much higher participation compared to Task C. This task involves classifying terms into predefined categories, a more familiar task for many researchers. Term typing is conceptually simpler because it involves assigning a label to a term, which is something that even general-purpose LLMs can do relatively well. There is a clear, finite set of categories or types, and many participants experimented with text classification approaches. 
\section{Conclusion}
The 1st Large Language Models for Ontology Learning Challenge at ISWC 2024 has revealed the emerging potential of LLMs beyond previous studies of OL tasks. The diverse range of participant systems, including fine-tuning, prompt-tuning, and retrieval-augmented generation approaches, demonstrated how adaptable LLMs can be when handling complex ontological data across various domains. The integration of diverse LLMs like GPT-4o, GPT-3.5, LLaMA-3, and Mistral underscored the versatility of LLMs. 

Through this challenge, key insights were garnered regarding the strengths and limitations of current LLMs for OL. Notably, while LLMs have shown a remarkable capacity to generalize across unseen tasks (as evidenced by their performance in few-shot and zero-shot scenarios), certain domains such as biomedical and geographical ontologies posed unique challenges, particularly in terms of class imbalance and complex taxonomies. These challenges opened pathways for future research, emphasizing the need for scalable LLM training and the refinement of prompt-based methods to handle highly specialized ontologies.

Moreover, the variety of approaches suggests that hybrid methods combining LLMs with domain-specific knowledge are particularly effective. Moving forward, research should focus on improving the interpretability and scalability of LLM-based OL systems to enable even more accurate and dynamic knowledge extraction. This challenge has laid the groundwork for expanding LLM capabilities in the context of the Semantic Web, fostering innovation and collaboration in building the next generation of intelligent web technologies.
\section*{Acknowledgements}
The 1st LLMs4OL Challenge @ ISWC 2024 jointly supported by the \href{https://www.nfdi4datascience.de/}{NFDI4DataScience initiative} (DFG, German Research Foundation, Grant ID: 460234259) and the \href{https://scinext-project.github.io/}{SCINEXT project} (BMBF, German Federal Ministry of Education and Research, Grant ID: 01lS22070).
%
%

\bibliographystyle{splncs04}
\bibliography{bibliography}
%




\end{document}